\documentclass[runningheads]{llncs}

\usepackage{graphicx}
\usepackage{subfigure}

\usepackage{xcolor}
\usepackage{amsfonts}
\usepackage{xspace}
\usepackage{array}
\usepackage{eucal}
\usepackage[algo2e,ruled,vlined,linesnumbered]{algorithm2e}
\SetKwRepeat{Do}{do}{while}

\usepackage{amsmath}
 
\newcommand{\revise}[1]{{#1}}

\usepackage{orcidlink}
\renewcommand{\orcidID}[1]{\raisebox{1.1ex}{\orcidlink{#1}}}

\begin{document}

\title{Non-Elitist Selection Can Improve the Performance of Irace}
\titlerunning{Non-Elitist Selection Can Improve the Performance of Irace}
\author{Furong Ye\inst{1}\orcidID{0000-0002-8707-4189}
\and Diederick Vermetten\inst{1}\orcidID{0000-0003-3040-7162} 
\and Carola Doerr\inst{2}\orcidID{0000-0002-4981-3227} 
\and Thomas B\"ack\inst{1}\orcidID{0000-0001-6768-1478}}

\institute{
LIACS, Leiden University, Leiden, The Netherlands 
\email{\{f.ye,d.l.vermetten,t.h.w.baeck\}@liacs.leidenuniv.nl} 
\and
Sorbonne Universit\'e, CNRS, LIP6, Paris, France
\email{carola.doerr@lip6.fr} 
}
\begin{sloppypar}
\maketitle

\begin{abstract} 
Modern optimization strategies such as evolutionary algorithms, ant colony algorithms, Bayesian optimization techniques, etc.~come with several parameters that steer their behavior during the optimization process. 
To obtain high-performing algorithm instances, automated algorithm configuration techniques have been developed. 
One of the most popular tools is irace, which evaluates configurations in sequential races, making use of iterated statistical tests to discard poorly performing configurations. At the end of the race, a set of \emph{elite} configurations are selected from those \emph{survivor} configurations that were not discarded, using greedy truncation selection.

We study two alternative selection methods: one keeps the best survivor and selects the remaining  configurations uniformly at random from the set of survivors, while the other applies entropy to maximize the diversity of the elites.
These methods are tested for tuning ant colony optimization algorithms for traveling salesperson problems and the quadratic assignment problem and tuning an exact tree search solver for satisfiability problems. The experimental results show improvement on the tested benchmarks compared to the default selection of irace.
In addition, the obtained results indicate that non-elitist can obtain diverse algorithm configurations, which encourages us to explore a wider range of solutions to understand the behavior of algorithms.

\end{abstract}


\section{Introduction}
\label{sec:Intro}
Algorithm configuration (AC) addresses the issue of determining a well-performing parameter configuration for a given algorithm on a specific set of optimization problems.
Many techniques such as local search, Bayesian optimization, and racing methods have been proposed and applied to solve the AC problem. 
The corresponding software packages, such as ParamILS~\cite{ParamILS}, SMAC~\cite{SMAC}, SPOT~\cite{SPOT},  MIP-EGO~\cite{wang2017new}, and irace~\cite{Iracepaper} have been applied to problem domains such as combinatorial optimization~\cite{Iracepaper}, software engineering~\cite{basmer2019encoding}, and machine learning~\cite{kotthoff2019auto}.

Irace, one of the most popular tools, has shown its ability to improve the performance of the algorithms for various optimization problems~\cite{SPEAR,CintranoFLA,Iracepaper,ACOTSP}.
However, we can still intuitively expect to improve the performance of irace considering contemporary optimization techniques.
Premature convergence is a common problem for optimization methods resulting in being trapped into local optima, which can also present irace from finding the optimal configurations.
For example, irace fails to find the optimal configuration of a family of genetic algorithms (GAs) for \textsc{OneMax} in~\cite{ye2021automated}. There exists more than one type of competitive configuration of the GA for \textsc{OneMax}, which is known due to the extensive body of theoretical work~\cite{GiessenW17Algorithmica,Sudholt12}. However,
irace converges to a specific subset of configurations that share similar algorithm characteristics.
In order to avoid issues like this, one could aim to increase the exploration capabilities of irace. 
However, this does not necessarily address the concern of finding well-performing configurations located in different parts of the space. 
Instead, we would want to allow irace to automatically explore search space around a diverse set of well-performing configurations to avoid converging on one specific type of configuration.

A ``soft-restart'' mechanism has been introduced for irace to avoid premature convergence in~\cite{Iracepaper}, which partially reinitializes the sampling distribution for the configurations that are almost identical to others.
However, evaluations can be wasted on testing similar configurations before the restart, and the configuration may converge on
\revise{the type of configurations that were found before restarting.}
Therefore, we investigate alternative selection mechanisms which take into account the diversity of the selected elite configurations.
In addition, the observations from~\cite{ye2021automated} inspire a discussion on searching for various competitive configurations with different patterns, which is addressed by our discussion that more knowledge can be obtained by searching diverse configurations. 

\subsection{Our Contributions}
In this paper, we show that an alternative random selection of elites can result in performance benefits over the default selection mechanism in irace. 
Moreover, we propose a selection operator maximizing the entropy of the selected elites. These alternative selection operators are compared to default irace on the tested scenarios.

The alternative approaches are tested on three scenarios:
tuning \revise{the Ant Colony Optimization (ACO) algorithm for} the traveling salesperson problem (TSP) and the quadratic assignment problem (QAP) and minimizing the computational cost of the SPEAR tool (an exact tree search solver for the satisfiability (SAT) problem). 

Experimental results show that (1) randomly selecting elites among configurations that survived the racing procedure
performs better than the greedy truncation selection, and (2) the irace variant that uses the entropy metric obtains diverse configurations and outperforms the other approaches. 
Finally, the obtained configurations encourage us to (3) use such a diversity-enhancing approach to find better configurations and understand the relationship between parameter settings and algorithm behavior for future work.

\textbf{Reproducibility:} 
\revise{We provide the full set of logs from the experiments described in this paper in~\cite{dataNonElitist}.} Additionally, our implementation of the modified irace versions described in this paper is available at~\url{https://github.com/FurongYe/irace-1}.

\section{Related work}
\label{sec:related}
\subsection{Algorithm Configuration}
Traditionally, the AC problem, as defined below, aims at finding a \emph{single} optimal configuration for solving a set of problem instances~\cite{eggensperger2019pitfalls}. 
\begin{definition}[Algorithm Configuration Problem] 
Given a set of problem instances $\Pi$, a parametrized algorithm $A$ with parameter configuration space $\Theta$, and a cost metric $c:\Theta \times \Pi \rightarrow \mathbb{R}$ that is subject to minimization, the objective of the AC problem is to find a configuration $\theta^* \in \underset{\theta\in\Theta}{\arg \min } \; \underset{\pi \in \Pi}{\sum} c(\theta,\pi)$.
\label{def:AC}
\end{definition}
The parameter space can be continuous, integer, categorical, or mixed-integer. 
In addition, some parameters can be conditional.

Many configurators have been proposed for the AC problem~\cite{SPOT,SMAC,ParamILS,li2017hyperband,Iracepaper,wang2017new}, and they usually follow Definition~\ref{def:AC} by searching for a \emph{single} optimal solution, although the solvers may apply population-based methods.
However, in some cases it can be desirable to find a set of diverse, well-performing solutions to the AC problem.
For example, previous studies~\cite{lopez2014automatically,ye2021automated} found that algorithm configurators can obtain different results when tuning for different objectives (i.e., expected running time, best-found fitness, and anytime performance), which suggests that a bi- or multi-objective approach to algorithm configuration can be a promising research direction. For such multi-objective configuration tasks, having diverse populations of configurations is a necessity to understand the Pareto front.

\subsection{Diversity Optimization}
To address the objective of obtaining a set of diverse solutions, certain evolutionary algorithms have been designed specifically to converge to more than one solution in a single run. 
For example, the Niching Genetic Algorithms are applied for solving multimodular functions~\cite{cavicchio1970adaptive,goldberg1987genetic} and searching diverse solutions of association rules~\cite{MARTIN2016208}, chemical structures~\cite{de2014dynamic}, etc. 
Diversity optimization also addresses the problem of searching for multiple solutions. 
Quality-diversity optimization~\cite{cully2017quality} was introduced to aim for a collection of well-performing and diverse solutions. 
The method proposed in~\cite{cully2017quality} measures the quality of solutions based on their performance (i.e., quality) and distance to other solutions (i.e., novelty) dynamically. 
The novelty score of solutions is measured by the average distance of the $k$-nearest neighbors~\cite{lehman2011abandoning}. 
Also, to better understand the algorithm's behavior and possible solutions, feature-based diversity optimization was introduced for problem instance classification~\cite{gao2016feature}. 
A discrepancy-based diversity optimization was studied on evolving diverse sets of images and TSP instances~\cite{neumann2018discrepancy}. 
The approaches in both studies measure the solutions regarding their features instead of performance.
Unfortunately, the AC problem usually deals with a mixed-integer search space, which is often not considered in the methods described in this section.

\section{Irace}
\label{sec:irace}

In this section, we describe the outline of irace. Irace is an iterated racing method that has been applied for hyperparameter optimization problems in many domains. It samples configurations (i.e., hyperparameter values) from distributions that evolve along the configuration process. Iteratively, the generated configurations are tested across a set of instances and are selected based on a racing method. The racing is based on statistical tests on configurations' performance for each instance, and elite configurations are selected from the configurations surviving from the racing. The sampling distributions are updated after selection. The distributions from sampling hyperparameter values are independent unless specific conditions are defined. As a result, irace returns one or several elite configurations at the end of the configuration process.
\begin{algorithm2e}
\textbf{Input:} Problem instances $\Pi = \{\pi_1,\pi_2,\ldots\}$, parameter configuration space 
% $\Theta$
\revise{$X$}, cost metric $c$, and tuning budget $B$\;
Generate a set of $\Theta_1$ sampling from $X$ uniformly at random\;
$\Theta^{\text{elite}} = \text{Race}(\Theta_1,B_1)$\;
\While{The budget $B$ is not used out}{
$j = j+1$\;
$\Theta_j = \text{Sample}(X,\Theta^\text{elite})$\;
$\Theta^\text{elite} = \text{Race}(\Theta_j \cup \Theta^\text{elite},B_j)$\;
 
}
\textbf{Output:} $\Theta^\text{elite}$
\caption{Algorithm Outline of irace}
\label{alg:irace}
\end{algorithm2e}

Algorithm~\ref{alg:irace} presents the outline of irace~\cite{Iracepaper}. 
Irace determines the number of racing iterations $N^{\text{iter}} = \lfloor 2+ \log_{2}(N^{\text{param}})\rfloor$ before performing the race steps, where $N^{\text{param}}$ is the number of parameters. For each $\text{Race}(\Theta_j,B_j)$ step, the budget of the number of configuration evaluations $B_j=(B-B_{\text{used}})/(N^{\text{iter}}-j+1)$, where $B_{\text{used}}$ is the used budget, and $j = \{1,\ldots,N^{\text{iter}}\}$.
After sampling a set of new configurations in each iteration, $\text{Race}(\Theta,B)$ selects a set of elite configurations $\Theta^{\text{elite}}$ (elites).
New configurations are sampled based on the parent selected from elites $\Theta^\text{elite}$ and the corresponding self-adaptive distributions of hyperparameters. Specific strategies have been designed for different types (numerical and categorical) of parameters.

Each race starts with a set of configurations $\Theta_j$ and performs with a limited computation budget $B_j$. Precisely, each candidate configuration of $\Theta_j$ is evaluated on a single instance $\pi_i$, and the configurations that perform statistically worse than at least another one will be discarded after being evaluated on a number of instances. Note that the irace package provides multiple statistical test options for eliminating worse configurations such as \revise{the} F-test and \revise{the} t-test. The race terminates when the remaining budget is not enough for evaluating the surviving configurations on a new problem instance, or when $N^{\text{min}}$ or fewer configurations survived after the test. At the end of the race, $N_j^{\text{surv}}$ configurations 
% remain survival 
\revise{survive} and are ranked based on their performance. 
Irace selects $\min\{N^{\text{min}}, N_j^{\text{surv}}\}$ configurations with the best ranks to form $\Theta^\text{elite}$ for the next iteration. 
Note that irace applies here a \emph{greedy elitist mechanism}, and this is the essential step where our irace variants alter in this paper.

To avoid confusion, we note that an ``elitist iterated racing'' is described in the paper introducing the irace package~\cite{Iracepaper}.
The ``elitist'' there indicates preserving the best configurations found so far. The idea is to prevent ``elite'' configurations from being eliminated due to poor performance on specific problem instances during racing, and \revise{the best  surviving ``elite'' configurations are selected to form $\Theta^\text{elite}$ }.  We apply this ``elitist racing'' for our experiments in this paper, while 
% non-elitist alternative methods are proposed for the selection of elites.
\revise{the alternative methods select diverse surviving ``elite'' configurations instead of the best ones.}
\section{Random Survivor Selection}
\label{sec:rand}
To investigate the efficacy of the greedy truncation selection mechanism used by default within irace, we compare the baseline version of irace to a version of irace that uses a random selection process. In particular, we adopt the selection of elites by \emph{taking the best-performing configuration and randomly selecting the remaining} $ N_j^{\text{surv}} - N^{\text{min}} - 1$ distinct ones from the best $\sigma N^\text{min}$ surviving configurations when $N^\text{surv} \ge \sigma N^\text{min}$, for some $\sigma \ge 1$. The implementation of our variants is built on the default irace package~\cite{iracegithub}.

\subsection{Tuning Scenario: ACOTSP}
\label{sec:Exp:acotsp}
ACOTSP~\cite{ACOTSP} is a package implementing ACO for the symmetric TSP.
We apply irace variants in this paper to configure 11 parameters (three categorical, four continuous, and four integer variables) of ACO for lower solution costs (fitness). 
The experimental results reported in the following are from $20$ independent runs of each irace variant. Each run is assigned with a budget of $5,000$ runs of ACOTSP, and ACOTSP executes 20s of CPU-time per run following the suggestion in~\cite{Iracepaper}.
We set $\sigma N^{\text{min}} = N^{\text{surv}}$ indicating that the irace variant, irace-rand, randomly selects survivor configurations to form elites. Other settings remain as default: the ``elitist iterated racing'' is applied, and $N^\text{min} =5$. We apply the benchmark set of Euclidean TSP instances of size $2,000$ with $200$ train and $200$ test instances.

\begin{figure}[!htb]
\vspace{-1mm}
    \centering
    \includegraphics[width=0.8\linewidth]{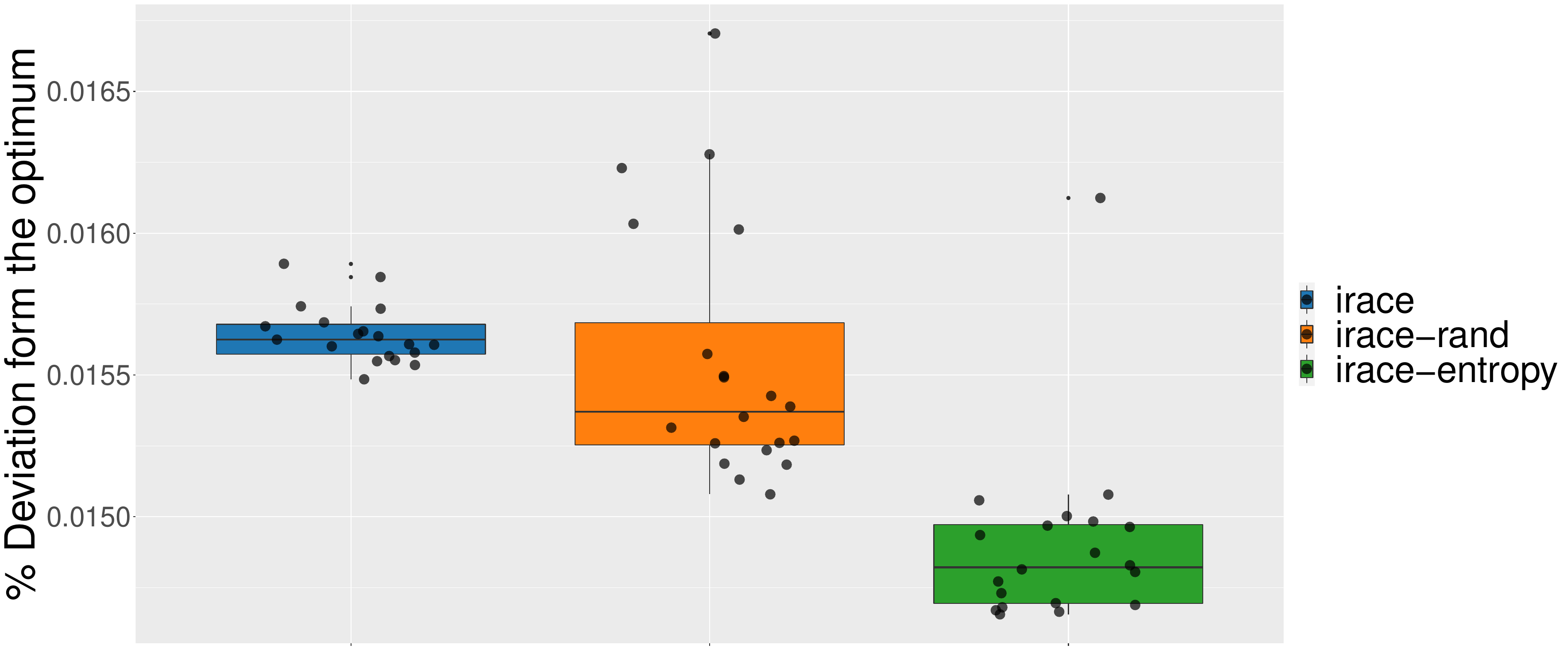}
    \caption{Average deviation from the optimum of the best obtained configurations. Each dot corresponds to the best final elite obtained by a run of irace, which plots the average deviation from the best-found fitness across $200$ TSP instances. Configurations are measured by the average result of $10$ validation runs per instance. The ``optimum'' for each instance is the best-found configuration obtained by $20$ ($60$ in total) runs of the plotted methods.}
    \label{fig:ACOTSP-method}
\end{figure}

Fig.~\ref{fig:ACOTSP-method} plots the deviations of the best configurations, which are obtained by each run, from the best-found (optimum) configuration obtained by $20$ ($60$ in total) runs of the irace variants. The results are averaged across $200$ TSP instances.
We observe that the median and mean of irace-rand results are smaller than those of irace, but the performance variance among these $20$ irace-rand runs is significantly larger.

\begin{figure}[!htb]
    \centering
    \includegraphics[width=0.8\linewidth]{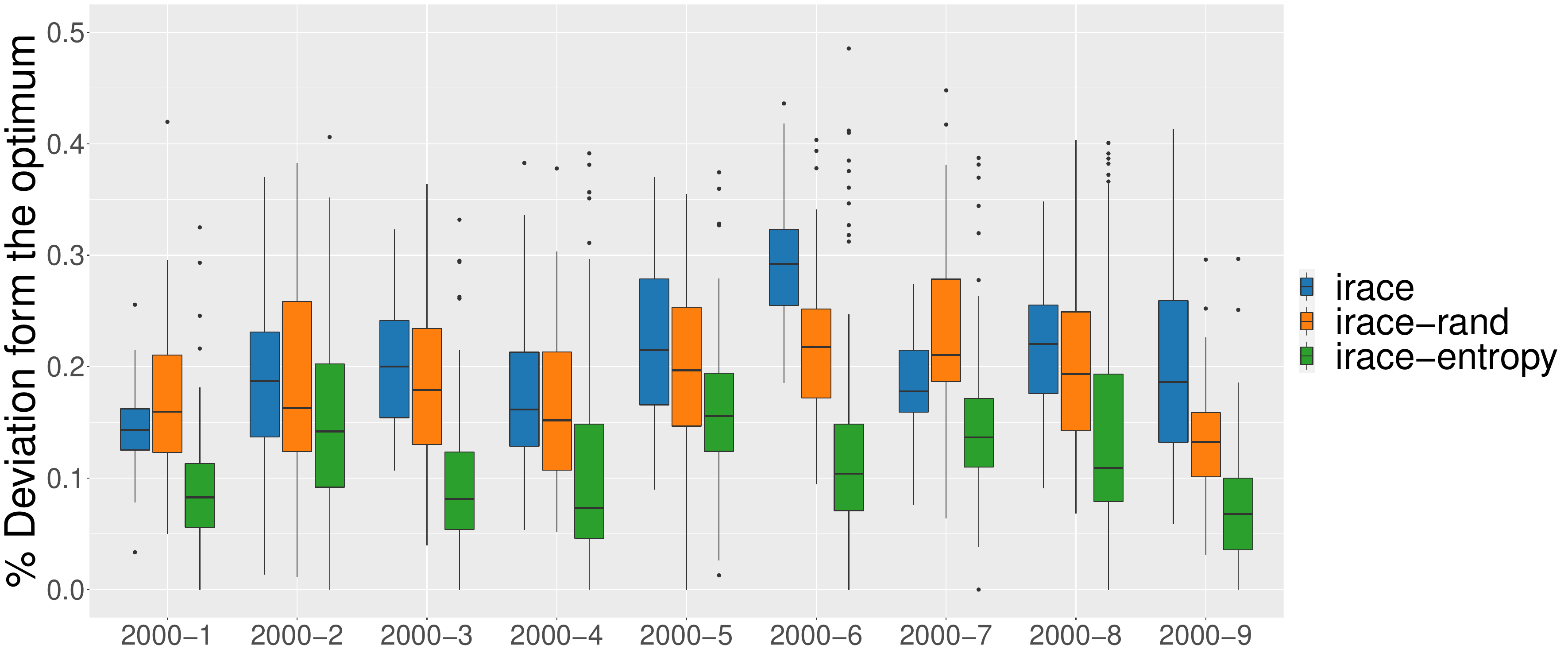}
    \caption{Boxplots of the deviation from the optimum of the obtained configurations for TSP instances. Results are from the average fitness of 10 validation runs for each obtained configuration.}
    \label{fig:ACOTSP-config}
\end{figure}

Though irace is initially proposed for searching configurations that generally perform well across a whole set of problem instances, we are nevertheless interested in the performance of the obtained configurations on individual instances. 
Therefore, we plot in Fig.~\ref{fig:ACOTSP-config} the performance of all obtained configurations on \revise{nine instances}.
Still, we observe comparable performance between irace and irace-rand.  
It is not surprising that the performance of irace-rand presents larger variance because the configurations that do not perform the best get a chance to be selected.
Moreover, we spot significant improvement on instances ``2000-6'' and ``2000-9'', on which the configurations obtained by irace-rand generally perform closer to the optima, compared to irace. 

\begin{figure}[htb]
    \centering
    \includegraphics[width=0.8\linewidth]{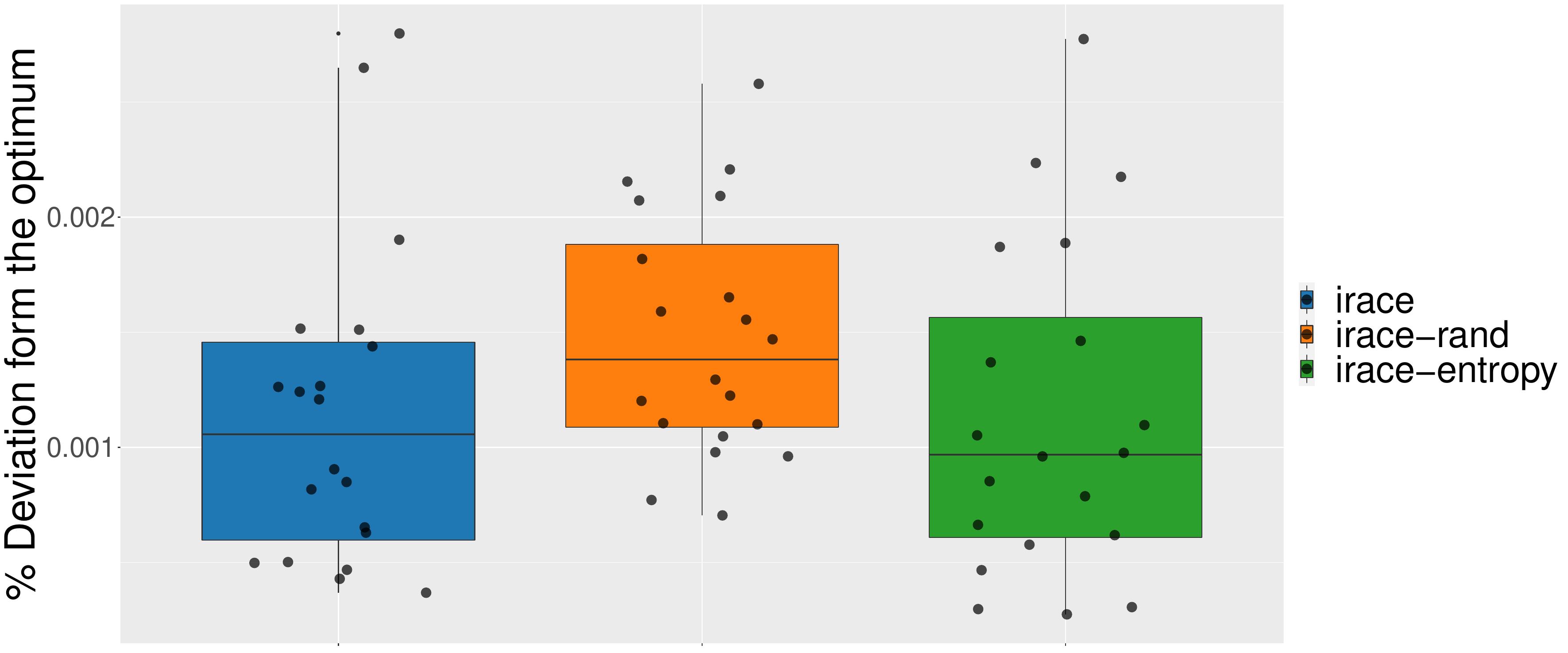}
    \caption{Average deviation from the optimum of the best obtained configurations. Each dot corresponds to the best final elite obtained by a run of irace, which plots the average deviation from the best-found fitness across $50$ QAP instances. Configurations are measured by the average result of $10$ validation runs per instance. The ``optimum'' is the best-found configuration obtained by $20$ ($60$ in total) runs of the plotted methods.}
    \label{fig:ACOQAP-method}
\end{figure}

\begin{figure}[htb]
    \centering
    \includegraphics[width=0.8\linewidth]{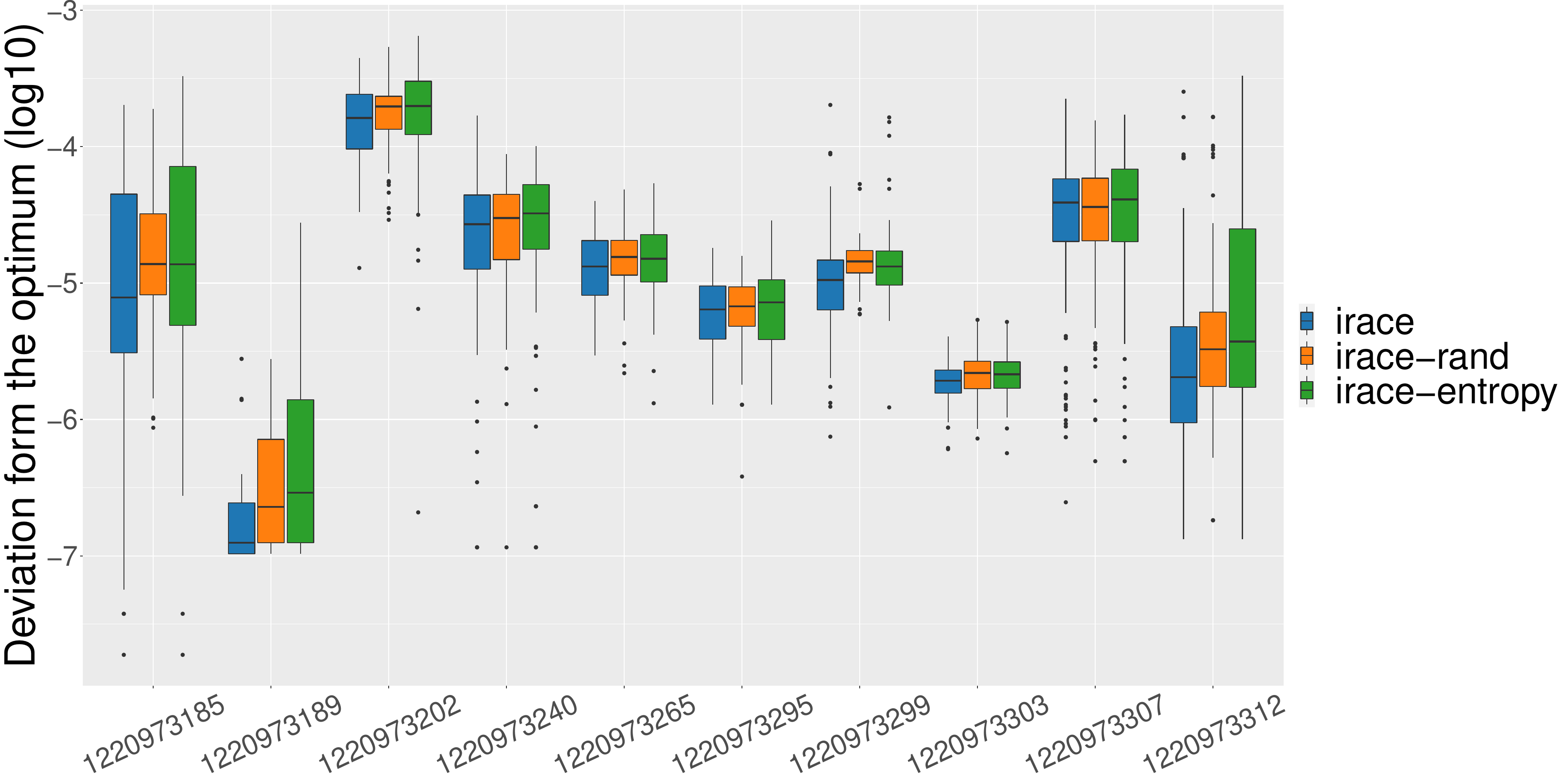}
    \caption{Boxplots of the deviation from the optimum of the obtained configurations for ACOQAP instances. Results are from the average fitness of 10 validation runs for each obtained configuration.}
    \label{fig:ACOQAP-Compare}
\end{figure}

\subsection{Tuning Scenario: ACOQAP} 
\revise{We apply the same ACO implementation in~\cite{Iracepaper} for solving QAP~\cite{lopez2015ant}.} ACOQAP executes 60s CPU-time per run following the default setting of the package, and we apply the benchmark set of $50$ train and test instances, respectively. The other settings remain the same with the ACOTSP scenario.

Unfortunately, we do not observe similar improvement of using irace-rand for ACOQAP. Irace-rand present worse performance than irace, comparing the average results across $50$ instances in Fig.~\ref{fig:ACOQAP-method}.
While looking at Fig.~\ref{fig:ACOQAP-Compare}, which plots the results on ten randomly picked instances, we do not observe improvement using irace-rand on ACOQAP, either.
These observations indicate that using this random selection to select elite configurations may deteriorate the performance of irace for ACOQAP, though it does not necessarily mean that diverse configurations are not helpful for the configuring process.
We will discuss this topic in more detail in Section~\ref{sec:diversity}.

\begin{figure}[htb]
    \centering
    \includegraphics[width=0.8\linewidth]{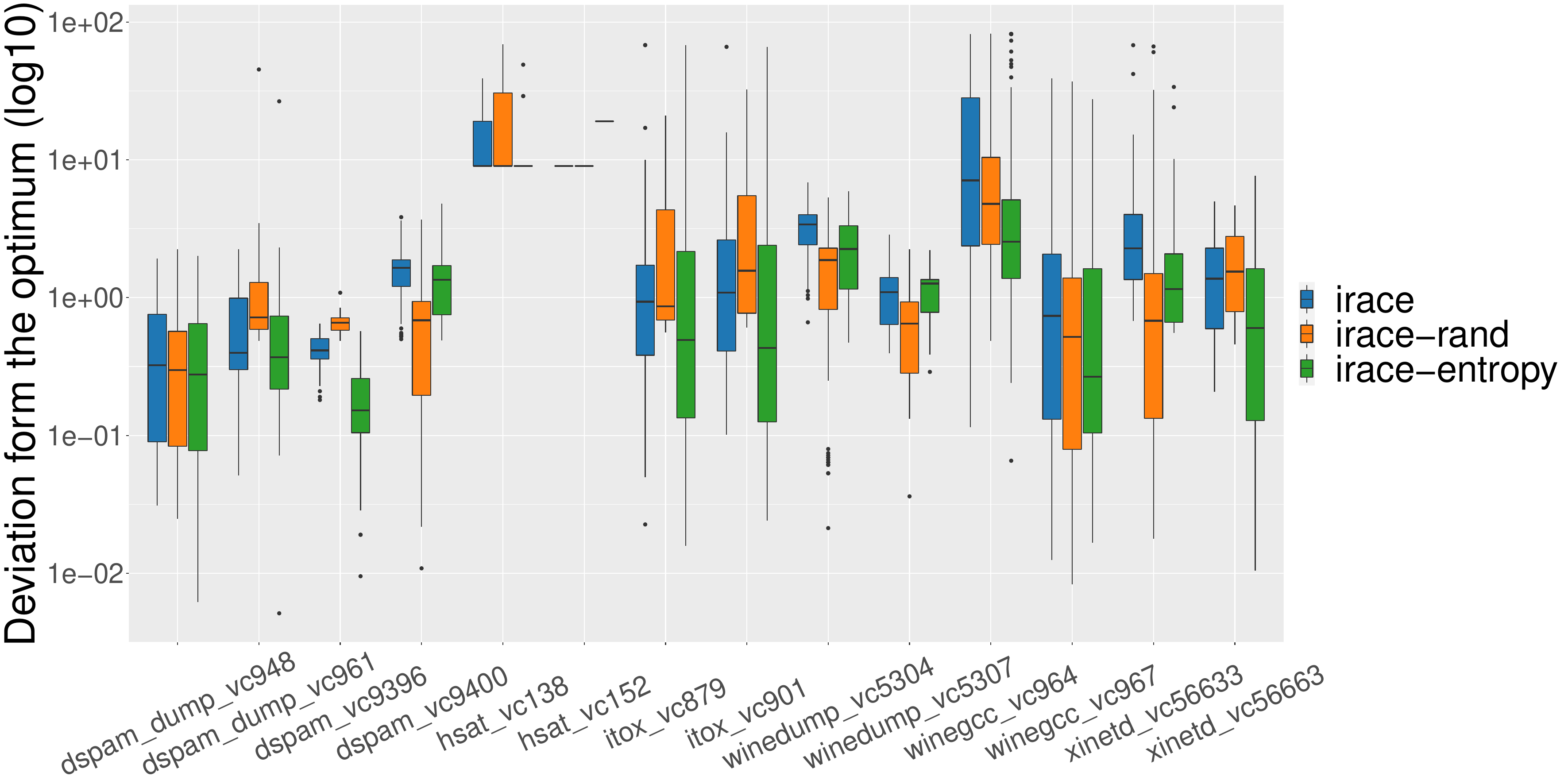}
    \caption{Boxplots of the deviation from the optimum of the obtained configurations for SPEAR instances. Results are from the average fitness of 10 validation runs for each obtained configuration. Results of ``gzip'' class is omitted because runtime of all obtained configurations are identical.}
    \label{fig:SPEAR-method}
\end{figure}

\subsection{Tuning Scenario: SPEAR}
SPEAR~\cite{SPEAR} is a custom-made SAT solver configurable with $26$ categorical parameters, of which nine are conditional, i.e., their activation depends on the values of one or several of the other parameters. Our goal here is to minimize the runtime of SPEAR. We run each irace variant $20$ independent times. Each run of irace is assigned with a budget of $10\,000$ runs of SPEAR, and the maximal runtime of SPEAR is 30s CPU-time per run. Other irace settings remain default: the ``elitist iterated racing'' is applied, and $N^\text{min} = 6$. The training and test set are $302$ different SAT instances, respectively~\cite{babic2007structural}.
Note that the number of survivor configurations is large ($\sim$250) during racing, and experimental results show that randomly selecting with such a large population deteriorates the performance of irace. Therefore, for this scenario, we cap the size of survivor candidates by $2N^\text{min} (\sigma = 2)$ to select from a relatively well-performing population.

Overall, we observe that the performance difference between the two methods is tiny for most instances, \revise{though irace-rand can not obtain better average results of runtime across all tested instances than irace}. Note that the obtained configurations may use much runtime ($\sim$30s) for a few instances, resulting in the comparison among the average runtime ($\sim$3s) across all instances can be significantly affected by the results on those particular instances.
Therefore, we plot only the runtime for the first two instances of each class of instances in Fig.~\ref{fig:SPEAR-method}. Compared to irace, though the performance of irace-rand deteriorates on ``itox'' instances, significant improvements using irace-rand can be observed on more instances such as ``dspam\_vc9400'', ``winedump'' instances, and ``xinetd\_vc56633''.

\section{Selecting Diverse Elites}
\label{sec:diversity}
The optimistic results of ACOTSP and SPEAR scenarios introduced in Section~\ref{sec:rand} indicate that, while keeping the best configuration, randomly selecting from well-performing survivor configurations to form elites can have positive impacts on the performance of irace.
An intuitive explanation is that irace-rand allows exploring search space around those \emph{non-elitist} configurations to avoid premature convergence on specific types of configurations, which matches our expectation following the motivation introduced in Section~\ref{sec:Intro}.
However, the failure to achieve improvements for ACOQAP requires us to consider explicitly controlling the selected elite configurations' diversity. To this end, we study an alternative selection strategy based on entropy~\cite{bromiley2004shannon} as a diversity measure.

\subsection{Maximizing Population Entropy}
In information theory, entropy represents random variables' information and uncertainty level~\cite{shannon2001mathematical}. The larger the entropy, the more information the variables deliver, e.g., the more diverse the solutions are.
Our irace-entropy configurator makes use of this idea, by using the Shannon entropy as criterion for selecting survivor configurations to form elites.

For a random variable $X$ with distribution $P(X)$, the normalized entropy of $X$ is defined as:
$$
H(X) = \sum_{i=1}^nP(X_i)\log P(X_i) / \log (n) ,
$$

In this paper, we estimate the entropy of integer and categorical variables from the 
\revise{probability} of each value. For continuous variables, the values are discretized into bins, and entropy is estimated based on the counts of each bin. \revise{Precisely, }the domain of a continuous variable is equally divided into $n$ bins, where $n$ is the number of observations (i.e., configurations). Finally, we calculate the diversity level $D(\Theta)$ of a set of configuration $\Theta$ using the mean entropy across $p$ variables (i.e., parameters), which is defined as:
$$
D(\Theta) = \frac{\sum_{j=1}^{p} H(\Theta^j) }{ p }, \quad \Theta^j = \{\theta_1^{j},  \theta_2^j,\ldots,\theta_n^j \}
$$

We introduce a variant of irace (irace-entropy) maximizing $D(\Theta^{\text{elite}})$ for each race step. Recall that $N^{\text{surv}}$ configurations survive at the end of race, and the $N^\text{min}$ best-ranked configurations are selected to form $\Theta^{\text{elite}}$ in Algorithm~\ref{alg:irace} (line 7). Irace-entropy adapts this step by selecting a subset of configurations $\Theta$ with the maximal $D^*(\Theta)$, where $|\Theta| = N^\text{min}$ and the best-ranked configuration $\theta^* \in \Theta$. In practice, we replace the greedy truncation selection in Algorithm~\ref{alg:irace} (line 7) with Algorithm~\ref{alg:maxEntropy}. \revise{Note that we do not explicitly handle conditional parameters.}

\begin{algorithm2e}
\textbf{Input:} A set of ranked configurations $\Theta^{\text{surv}}$, the maximal size $N^{\text{min}}$ of $\Theta^\text{elite}$ \;
\eIf{$|\Theta^{\text{surv}}| \le N^{\text{min}}$}{
$\Theta^\text{elite} = \Theta^{\text{surv}}$}{
$\Theta^\text{elite} = \{\theta^*\}$, $\Theta^\text{surv} = \Theta^\text{surv} \backslash \{\theta^*$\}, where $\theta^* \in \Theta^{\text{surv}}$ is the best-ranked\;
$\Theta^\text{elite} = \Theta^\text{elite} \cup S^*$, where $ S^* = \underset{S \subset \Theta^\text{surv},  |S|=N^{\text{min}}-1}{\arg \max} D (\Theta^\text{elite} \cup S)$}
\textbf{Output:} $\Theta^\text{elite}$
\caption{Entropy-maximization selection}
\label{alg:maxEntropy}
\end{algorithm2e}

\subsection{Experimental Results}

We present the results of irace-entropy in this section. 
All the settings remain the same as reported in Section~\ref{sec:rand} while applying the introduced alternative selection method.

For ACOTSP, we observe in Fig.~\ref{fig:ACOTSP-method} that irace-entropy performs better than irace and irace-rand, obtaining significantly smaller deviations from the optimum than those of irace for $19$ out of $20$ runs.
Regarding the results on individual problem instances, irace-entropy also shows in Fig.~\ref{fig:ACOTSP-config} significant advantages against irace and irace-rand across all the plotted instances.

Recall that, for ACOQAP, the performance of irace-rand deteriorates compared to irace by randomly selecting survivor configurations to form elites. However, through using entropy as the metric to control the diversity explicitly, irace-entropy shows comparable results to irace in Fig.~\ref{fig:ACOQAP-method}, obtaining a smaller median of deviations from the optimum for $20$ runs. We also observe that the performance of irace-entropy is comparable to irace for individual instances in Fig.~\ref{fig:ACOQAP-Compare}. In addition, irace-entropy can obtain the best-found configurations for some instances such as ``1220973202'' and ``1220973265''.

For SPEAR, we observe in Fig.~\ref{fig:SPEAR-method} that irace-entropy outperforms irace. For $12$ out of the $14$ plotted instances, irace-entropy obtains better median results than irace. Moreover, irace-entropy achieves improvements compared to irace-rand for most instances. Especially for the ``itox'' instances, in which irace-rand does not perform as well as irace, irace-entropy obtains better results while also keeping the advantages over irace on other instances.

According to these results, we conclude that non-elitist selection can help improve the performance of irace. By using entropy as the metric to maximize the diversity of the selected elite configurations, irace-entropy achieves improvements compared to irace. However, irace-entropy does not obtain significant advantages against irace for ACOQAP and performs worse than irace-rand on some SPEAR instances, indicating potential improvements for non-elitist selection through further enhancements in regards to controlling the diversity of elites for specific problem instances.

\begin{figure*}[htb]
    \centering
    \subfigure[irace]{
         \includegraphics[trim=0 30 0 10, clip, width=.47\linewidth]{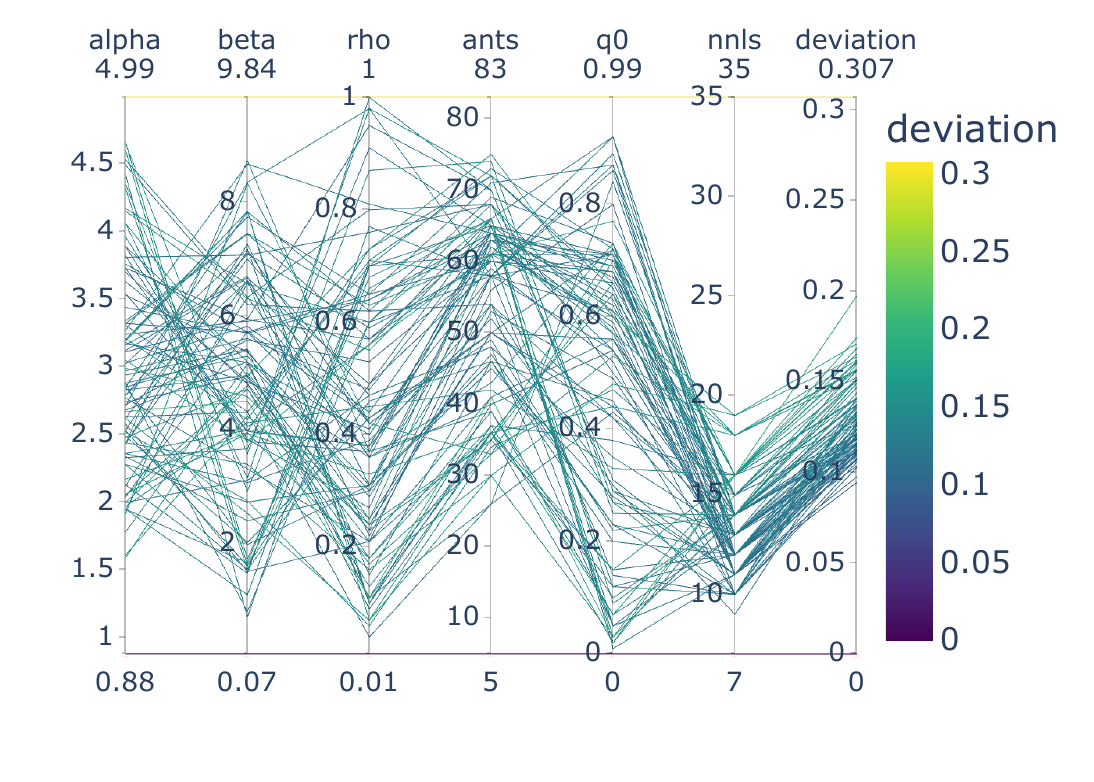}
    }
    \subfigure[irace-rand]{
         \includegraphics[trim=0 30 0 10, clip, width=.47\linewidth]{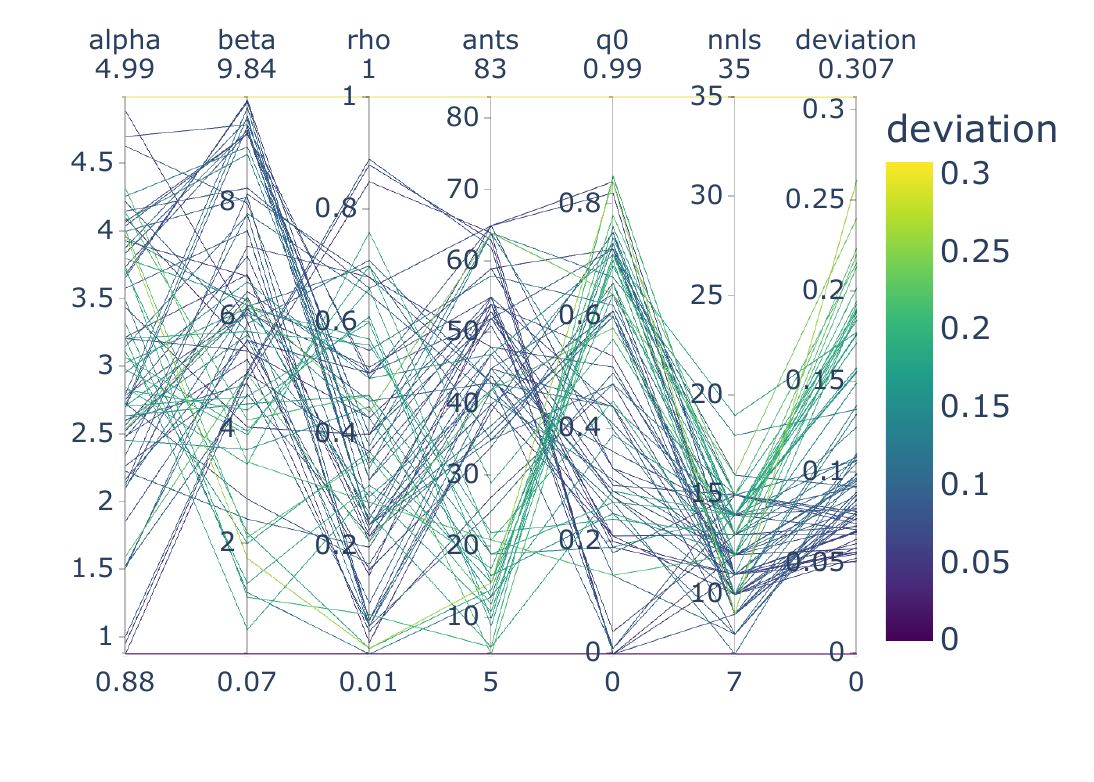}
    }
    \subfigure[irace-entropy]{
         \includegraphics[trim=0 30 0 10, clip,width=.47\linewidth]{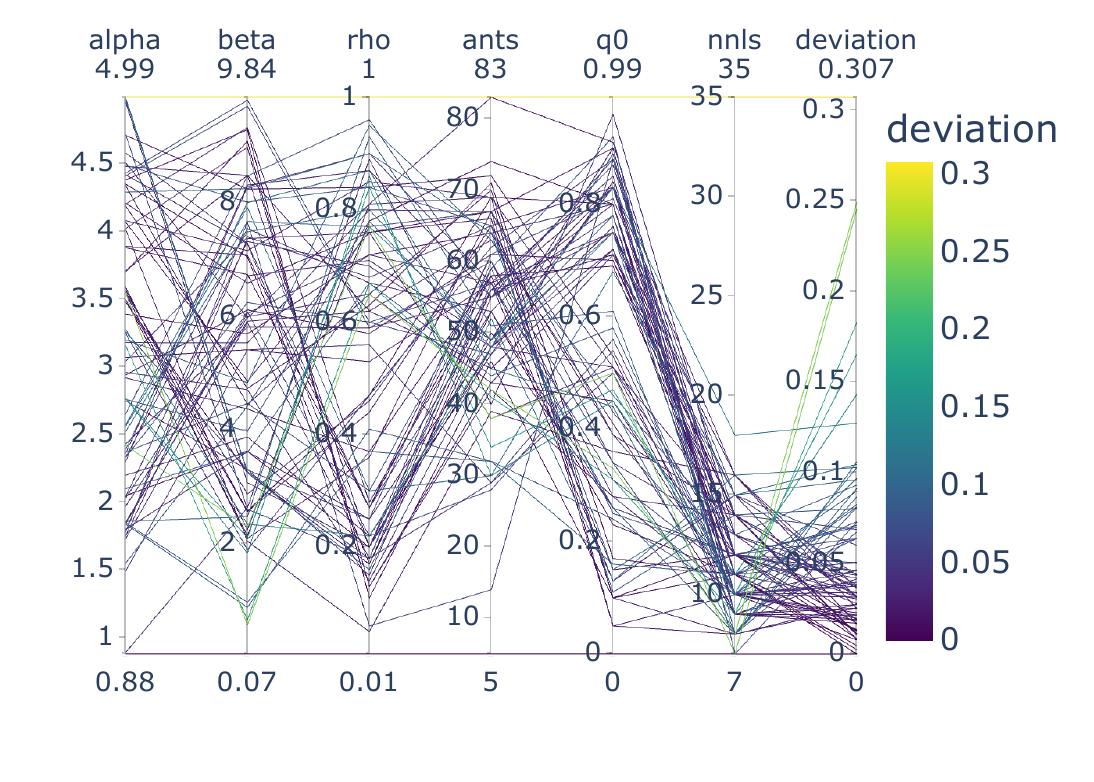}
     }
    \caption{Parameter values and deviations from the optimum of the ACOTSP configurations obtained by the irace variants. Each configuration is represented as a polyline with vertices on the parallel axes. The point of the vertex on each x-axis corresponds to the parameter value. We process ``Null'' values of the conditional parameter as~$0$. The color of the lines indicates the deviation of their average solution costs across all tested instances from that of the best-found one. Darker lines indicate better configurations.}
    \vspace{-0.2cm}
    \label{fig:PC}
\end{figure*}

\subsection{Benefits from Diverse Configurations}
\label{sec:div:learn}
While the irace variants, i.e., irace-rand and irace-entropy, achieve improvements by using non-elitist selection, significant variances are noticeable in Figures~\ref{fig:ACOTSP-config},~\ref{fig:ACOQAP-Compare}, and~\ref{fig:SPEAR-method} for results of the obtained configurations. Recall that AC techniques have been applied in~\cite{ye2021automated} for exploring promising configurations of the GA on diverse problems. Apart from analyzing a single optimal configuration, such benchmarking studies can also benefit from diverse configurations to investigate algorithms' performance with specific parameter settings. Therefore, we illustrate in this section that non-elitist selection can not only improve the performance of irace but also help understand the behavior of algorithms.

Using ACOTSP as an example, we show the configurations obtained by each irace variant in Fig.~\ref{fig:PC}.
The color of the configuration lines are scaled by the deviation $\frac{f - f^*}{f^*} \%$ from the optimum, where $f$ is the average solution cost across $200$ problem instances.

We observe that irace-entropy obtains most of the competitive configurations while covering a wider range of performance with deviations from $0$ to $0.25$. 
However, the performance of the configurations obtained by the irace cluster in a range of deviations from $0.1$ to $0.2$.
Moreover, regarding the parameters of the obtained configurations, the range of $beta$ and $q0$ are narrower for irace compared to the other methods. However, the configurations with $beta > 8$ and $q0 > 0.9$, outside the range obtained by irace, generally perform well.

We will not investigate how the parameter values practically affect the performance of ACOTSP since it is beyond the scope of this paper. Nonetheless, Fig.~\ref{fig:PC} provides evidence that irace-entropy can provide more knowledge concerning the distribution of obtained performance (i.e., fitness) and parameter values, which is helpful for understanding algorithms' behavior.

\section{Conclusions and Discussions}
\label{sec:conclusions}
In this paper, we have demonstrated that randomly selecting survivor configurations can improve the performance of irace, as illustrated on the cases of
tuning ACO on TSP and tuning SPEAR to minimize the runtime of a SAT solver. Moreover, we have proposed an alternative selection method to form diverse elite configurations, using Shannon entropy as the diversity metric. Experimental results show significant advantages of maximizing entropy in this way. 

While the irace-entropy presents improvement in the performance of irace via exploring diverse configurations in all the tested scenarios, irace-rand obtain better configurations for specific SPEAR instances. Therefore, there is still room for further study of incorporating diversity into the selection operators. 
\revise{More in-depth analysis on a wider set of algorithm configuration problems can help us better understand the benefits of considering diversity in selection. }
In addition, we did not modify the procedure of sampling new configurations. Nevertheless, we believe effectively generating diverse configurations can be beneficial and shall be studied for future work.

Apart from boosting the performance of irace via focusing more on diversity, 
we can find a diverse portfolio of well-performing algorithm configurations while keeping the benefits of the iterated racing approach, by changing the objective of the tuning from finding the best performing configuration to find a diverse portfolio of well-performing algorithm configurations. 
In the context of algorithm selection, such approaches are studied under the notion of \emph{algorithm portfolio selection}~\cite{Lindauer2018PortfolioAS}. 

\end{sloppypar}

\bibliographystyle{splncs04}
\bibliography{reference}

\end{document}